\documentclass[10pt,twocolumn,letterpaper]{article}

\usepackage[pagenumbers]{cvpr} % To force page numbers, e.g. for an arXiv version

\usepackage{graphicx}
\usepackage{amsmath}
\usepackage{amssymb}
\usepackage{booktabs}
\usepackage{color}
\usepackage{bbm}
\usepackage{multirow}
\usepackage{float}
\floatstyle{plaintop}
\restylefloat{table}
\usepackage{placeins}
\usepackage{graphicx}
\usepackage{caption}
\usepackage{subcaption}
\usepackage{xcolor}
\definecolor{ao(english)}{rgb}{0.0, 0.5, 0.0}
\graphicspath{ {./images/} }

\usepackage[pagebackref,breaklinks,colorlinks]{hyperref}

\usepackage[capitalize]{cleveref}
\crefname{section}{Sec.}{Secs.}
\Crefname{section}{Section}{Sections}
\Crefname{table}{Table}{Tables}
\crefname{table}{Tab.}{Tabs.}

\def\cvprPaperID{*****} % *** Enter the CVPR Paper ID here
\def\confName{CVPR}
\def\confYear{2022}

\begin{document}

\title{MATrIX - Modality-Aware Transformer for Information eXtraction}

\author{Thomas Delteil*\\
AWS AI\\
{\tt\small tdelteil@amazon.com}
\and
Edouard Belval*\\
AWS AI\\
{\tt\small belvae@amazon.com}
\and
Lei Chen\\
AWS AI\\
{\tt\small chenzlei@amazon.com}
\and
Luis Goncalves\\
AWS AI\\
{\tt\small luisgonc@amazon.com}
\and
Vijay Mahadevan\\
AWS AI\\
{\tt\small vmahad@amazon.com}
}
\maketitle
\begin{abstract}
We present MATrIX - a Modality-Aware Transformer for Information eXtraction in the Visual Document Understanding (VDU) domain. VDU covers information extraction from visually rich documents such as forms, invoices, receipts, tables, graphs, presentations, or advertisements. In these, text semantics and visual information supplement each other to provide a global understanding of the document. MATrIX is pre-trained in an unsupervised way with specifically designed tasks that require the use of multi-modal information (spatial, visual, or textual). We consider the spatial and text modalities all at once in a single token set. To make the attention more flexible, we use a learned modality-aware relative bias in the attention mechanism to modulate the attention between the tokens of different modalities. We evaluate MATrIX on 3 different datasets each with strong baselines.
\end{abstract}
\let\thefootnote\relax\footnotetext{* indicates equal contribution}

\section{Introduction}

The worldwide number of digitized documents is in the trillions according to Adobe, the company that developed the PDF format. Some of these documents, like scanned books or reports, mostly contain pages of linear text. However, a large portion of these documents are visually rich, containing forms, tables, or where the layout itself conveys important clues about the meaning of the text. Automating form processing, filling databases, processing invoices, parsing census information, or normalizing form values across heterogeneous layouts require performing complex information extraction tasks such as:
\begin{itemize}
    \item Document classification
    \item Sequence labeling, which consists of extracting groups of words in a document and labeling them.
    \item Table detection, where one tries to extract the location and structure of tabular data embedded in a document such as utility bills or financial reports
    \item Form recognition, which consists of extracting key-value pairs in documents like names, phone numbers, or order numbers.
\end{itemize}
To obtain good performance on these tasks, using the text modality alone is not sufficient\cite{DBLP:journals/corr/abs-1912-13318}, especially when the reading order is non-trivial. Transformer models relying on self-attention mechanisms have become instruments of choice for modeling simply and efficiently multi-modal input as they operate on sets of undifferentiated tokens in their simplest form. Modality fusion is therefore the crux of the problem. Previous approaches such as LayoutLM\cite{DBLP:journals/corr/abs-1912-13318} have simply concatenated vision and text tokens together. DocFormer\cite{DBLP:journals/corr/abs-2106-11539} introduced a complex specifically designed attention layer that fuses text and vision modality by summing them together, with the caveat that text and vision modalities must have the same length. Our approach aims at striking a better trade-off in complexity and efficiency by introducing a new modality-aware relative attention mechanism that allows individual tokens to modulate the amount of attention afforded to other tokens, conditioned on their own and their target tokens modality and relative spatial information. We also introduce a novel pre-training task: line bounding box regression to incentivize word tokens from the same lines to attend to each other visually and semantically with respect to their spatial location. We depart from the other works by using whole word tokens instead of sub-word tokens for our text modality representation by using the average embedding of the sub-words for each word. This allows us to work with much longer text lengths. Another key difference is that we represent all spatial information in the $[0, 1]$ relative domain of the non-padded vision input instead of using absolute pixel coordinate embedding, which allows us to use arbitrary resolution at inference time.

\section{Related work}

Information Extraction (IE) is the task of extracting structured information from an unstructured source\cite{niklaus-etal-2018-survey}, with early approaches relying on pattern matching and natural language processing to identify information based on its position\cite{califf-mooney-1997-relational}\cite{Cardie_1997}. With the emergence of deep learning approaches in NLP\cite{10.1145/1390156.1390177}, Document Understanding (DU) and its associated sub-tasks: key information extraction (KIE)\cite{park2019cord}\cite{DBLP:journals/corr/abs-1905-13538}, document layout analysis (DLA)\cite{Antonacopoulos2009ARD}\cite{DBLP:journals/corr/abs-1908-07836}\cite{DBLP:journals/corr/abs-2006-01038}, and document question answering (DQA)\cite{DBLP:journals/corr/abs-2007-00398} took the forefront.\\
Visual Document Understanding (VDU) is a subfield of Document Understanding that combines features from text and image to extract information from structured documents. Yang et al.\cite{DBLP:journals/corr/YangYAKKG17} were the first to introduce an end-to-end multi-modal approach for semantic structure labeling using a convolutional neural network and pre-trained natural language processing models. Devlin et al.\cite{DBLP:journals/corr/abs-1810-04805} then introduced BERT and showcased the potential of masked language-model pre-training on NLP tasks. Lu et al.\cite{DBLP:journals/corr/abs-1908-02265} introduced a co-attention mechanism that allows for the joint representation of image and text. LayoutLM\cite{DBLP:journals/corr/abs-1912-13318} showed that a BERT\cite{DBLP:journals/corr/abs-1810-04805} architecture combined to embeddings from Faster R-CNN could leverage bigger datasets using pre-training tasks to achieve SOTA. Shortly after, LayoutLMv2\cite{DBLP:journals/corr/abs-2012-14740} improved upon LayoutLM by using a single pre-training framework and combining the visual features and the text tokens at the input of the model.
Powalski et al.\cite{DBLP:journals/corr/abs-2102-09550} introduced TILT, an encoder-decoder approach, inspired from T5\cite{DBLP:journals/corr/abs-1910-10683} which inputs the pairwise distances at the decoder level  as a spatial bias.\\ DocFormer\cite{DBLP:journals/corr/abs-2106-11539} introduced an attention layer to fuse visual, textual and spatial features, but also a formal definition for visual and textual feature fusion approaches.
Additionally, BROS\cite{DBLP:journals/corr/abs-2108-04539} also used a BERT architecture with 2D spatial embeddings but forgoes transformer-based decoders in favour of SPADE\cite{DBLP:journals/corr/abs-2005-00642} to model more complex spatial relationships.

\section{Approach}
    
\subsection{Model Architecture}

Modality-Aware Transformer for Information eXtraction or MATrIX for short is an end-to-end trained encoder-only transformer architecture that relies on a pre-trained backbone for visual feature extraction. Following the conceptual categories introduced by Appalaraju et al.\cite{DBLP:journals/corr/abs-2106-11539} our approach is a joint multi-modal architecture in which the vision and language features are concatenated before applying the attention-based fusion layers.

\begin{figure}
\begin{center}
    \includegraphics[scale=0.25]{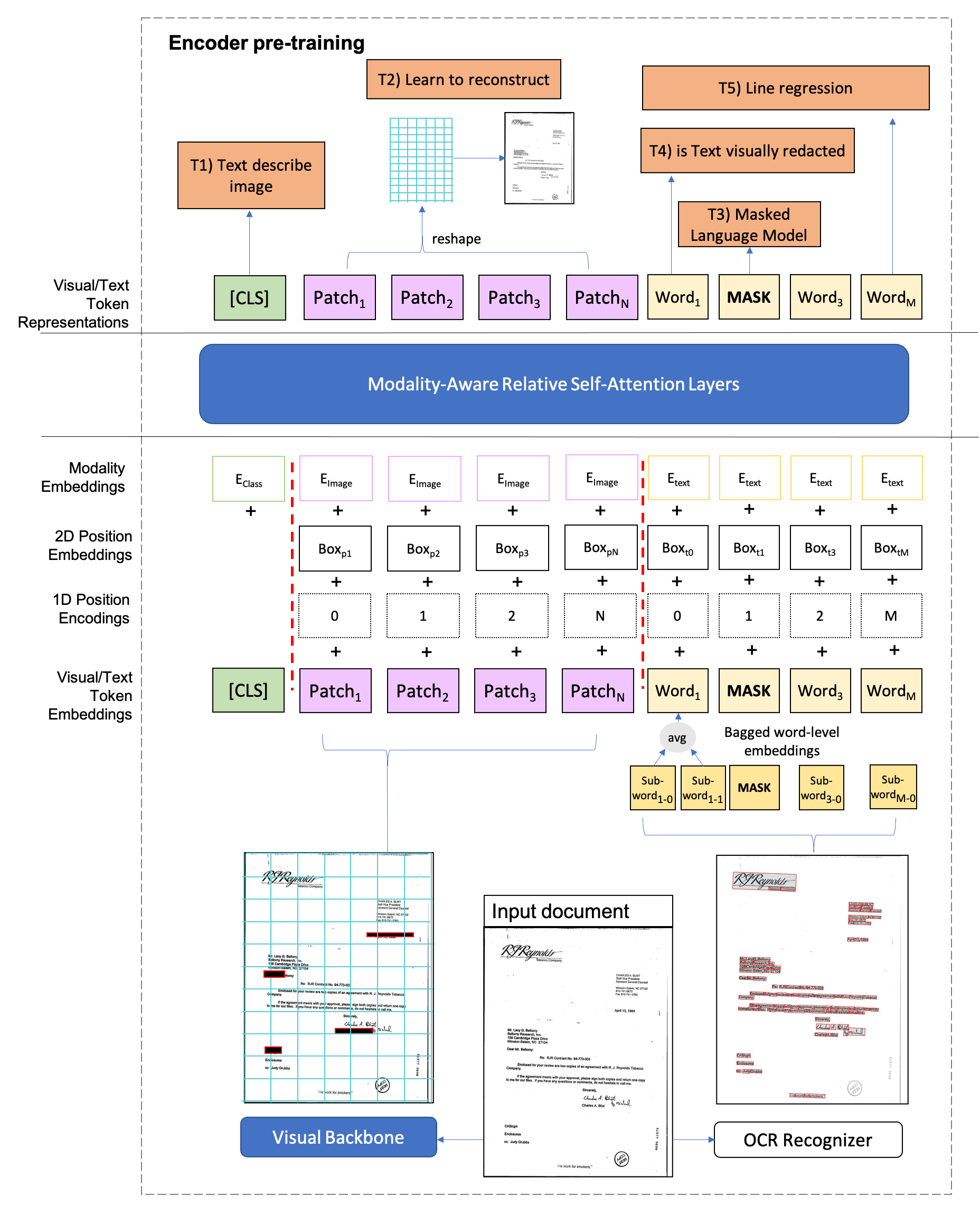}
    \caption{MATrIX pre-training high level overview}
    \label{fig:test}
\end{center}
\end{figure}

\subsubsection{Text features}

Following prior work\cite{DBLP:journals/corr/abs-1912-13318}\cite{DBLP:journals/corr/abs-2012-14740}\cite{DBLP:journals/corr/abs-2106-11539}, we tokenize the input words using WordPiece\cite{DBLP:journals/corr/WuSCLNMKCGMKSJL16}. Given a word $w$, the tokenizer $T$ will output $n$ tokens which are then passed through an embedding $W$, giving $n$ representations which are averaged at the word-level to create the encoder input text token $t$. Formally:

\begin{align}
T: 1\text{x}1 \longrightarrow n\text{x}1 \\
W: n\text{x}1 \longrightarrow n\text{x}n \\
t = \frac{1}{n}\sum_{i=1}^{n}W(T(w))
\end{align}

This approach allows us to train with a maximum sequence length of 512 word tokens as opposed to 512 sub-word tokens in previous work, allowing us to process entire, large documents with their entire text content.

\subsubsection{Visual features}

To enrich the document representation we add visual tokens to the input. Given an image, we resize the long side to 512 pixels, preserving the aspect ratio, we define a reduction ratio of 32 which gives us a maximum of 16x16 visual tokens. In a departure from previous work, we do not enforce a visual token sequence length and we can handle arbitrary resolutions given padding to the nearest reduction ratio multiple. To each vision token, we add a spatial embedding based on its feature map element size with respect to the non-padded image area. The weights of the visual-spatial embedding layer are shared with the text spatial embedding.\\

\subsubsection{Spatial features}

To each text and visual token, we add a spatial embedding to provide the encoder with 2D spatial information. As opposed to DocFormer which embedded the bounding box and LayoutLMv2 which used two embeddings, we encode the token index, top-corner $x$ and $y$ relative positions, and the relative width and height of the token. Formally $\text{pos} = (i, x, y, w, h)$. The embedding itself is a 3-layer feed-forward network with leaky ReLU activation.\\
We use the same embedding for visual and text tokens. The intuition behind this choice is that both types of features exist in the same 2D space.

\subsubsection{Relative Attention}

Vision and text features are usually heavily spatially correlated in VDU. In order to include this inductive bias, LayoutLMv2\cite{DBLP:journals/corr/abs-2012-14740} introduced a learn-able relative attention bias, learned independently for each attention head. 

\begin{align}
\alpha_{i j}^{\prime}=\alpha_{i j}+\mathbf{b}_{j-i}^{(1 \mathrm{D})}+\mathbf{b}_{x_{j}-x_{i}}^{\left(2 \mathrm{D}_{\mathrm{x}}\right)}+\mathbf{b}_{y_{j}-y_{i}}^{\left(2 \mathrm{D}_{\mathrm{y}}\right)}
\end{align}
This introduces a simple mechanism for the model to increase the self-attention of close-by tokens and reduce the attention of far-away tokens. We modified slightly this implementation to condition the relative attention bias jointly on the $x$ and the $y$ axis instead of two independent biases. 

\begin{align}
\alpha_{i j}^{\prime}=\alpha_{i j}+\mathbf{b}^{\left(2 \mathrm{D}_{\mathrm{x,y}}\right)}(x_{j}-x_{i},y_{j}-y_{i})
\end{align}

\subsubsection{Modality-Aware Relative Attention}

One of the difficult challenges for a multi-modal transformer is to build token representations that can be compared and combined across modalities through self-attention. Even if vision and text modalities both belong in the 2D space and share a common spatial representation, it doesn't mean that their spatial correlations are identical. For example, vision features might help the model uniformly around a given text token, while for text, the 1D ordering of a sentence that is over-represented in the $x$ axis relationship, is the most meaningful. We adapt the previously described relative attention bias by conditioning the learned bias based on the pair-wise modality of the tokens being considered. 

\begin{align}
\begin{split}
    \alpha_{i j}^{\prime} =\, & \alpha_{i j}\, +\, \\ 
    & \mathbf{b}^{\left(modality, 2 \mathrm{D}_{\mathrm{x,y}}\right)}(x_{j}-x_{i},y_{j}-y_{i}, mod_{i \longrightarrow j})
\end{split}
\end{align}

\subsection{Pre-training Tasks}

\begin{figure}
\centering
\begin{subfigure}{.23\textwidth}
  \centering
  \includegraphics[width=.9\linewidth]{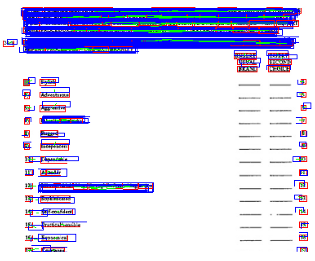}
  \caption{Line regression}
  \label{fig:sub1}
\end{subfigure}%
\begin{subfigure}{.23\textwidth}
  \centering
  \includegraphics[width=.9\linewidth]{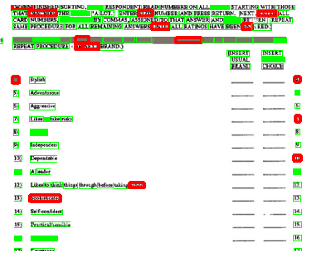}
  \caption{Token switch}
  \label{fig:sub2}
\end{subfigure}
\begin{subfigure}{.23\textwidth}
  \centering
  \includegraphics[width=.9\linewidth]{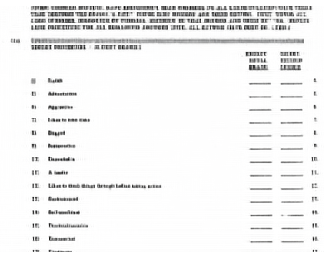}
  \caption{Reconstruct}
  \label{fig:sub2}
\end{subfigure}
\begin{subfigure}{.23\textwidth}
  \centering
  \includegraphics[width=.9\linewidth]{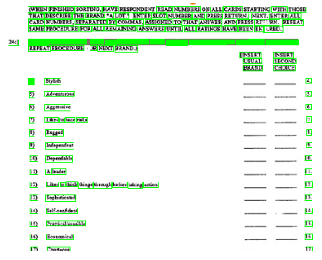}
  \caption{Line redaction}
  \label{fig:sub2}
\end{subfigure}
\caption{Visualizations of four pre-training tasks}
\label{fig:test}
\end{figure}

\subsubsection{Line Regression}

In this novel task, we infer the line bounding box of the input text tokens. Given the output of the encoder, we apply a linear projection for each text token to predict its line coordinates as $(x, y, w, h)$. The concept of line is defined as returned by the OCR engine which is a group of words separated by spaces. We use a mean-square error loss denoted $L_{LReg}$.

\begin{align}
L_{LReg} = \sum_{i \in [1,2]}^{W}(x_i- \hat x_i)^2 + (y_i- \hat y_i)^2
\end{align}
The model can achieve this objective only by learning to group words together semantically and visually to combine their spatial information in order to output the correct line for each word in a given line. 

\subsubsection{Token Switch}

In this novel VDU task, we switch text tokens in the input and use a projection to predict whether a token was switched. We use this task to encourage the model to learn a richer semantic understanding of the text sequence. We optimize the objective using a cross-entropy loss represented by $L_{TS}$.

% This is quite big, prefer \textbf?
\subsubsection{Multi-Modal Masked Language Modeling}

In this task, we try to predict masked text tokens. We mask or replace tokens in the original sequence $s$ and use the encoder to predict the correct token $t$ at a given position. In a similar fashion to DocFormer\cite{DBLP:journals/corr/abs-2106-11539}, we use a modified formulation of the original BERT\cite{DBLP:journals/corr/abs-1810-04805} MM-MLM in which the visual region is not masked, as opposed to prior work by Xu et al.\cite{DBLP:journals/corr/abs-1912-13318}\cite{DBLP:journals/corr/abs-2012-14740}. Our implementation differs from DocFormer as it accounts for the mean bag that is applied to the output of the tokenizer. We apply an L1 loss between the input and the output text embeddings, however, this approach is prone to collapsing as the embedding converges to a null vector $\Vec{0}$. To prevent this side-effect, we rely on the token switch task which forces the model to converge towards a solution that also allows for the differentiation of the text embeddings and thus makes the $\Vec{0}$ solution invalid.

\subsubsection{Learn To Reconstruct}

In this task, introduced by Appalaraju et al.\cite{DBLP:journals/corr/abs-2106-11539} we seek to reconstruct the input image from the multi-modal representation generated by the encoder. We use a simplistic decoder based on convolutional and up-sampling layers to infer the input image. We compute the error using a smooth L1 loss denoted as $L_{LTR}$. This task is expected to force the model to leverage textual features to reconstruct characters from the original image.

\begin{figure*}[h]
\centering
\begin{subfigure}{.5\textwidth}
  \centering
  \includegraphics[width=.6\linewidth]{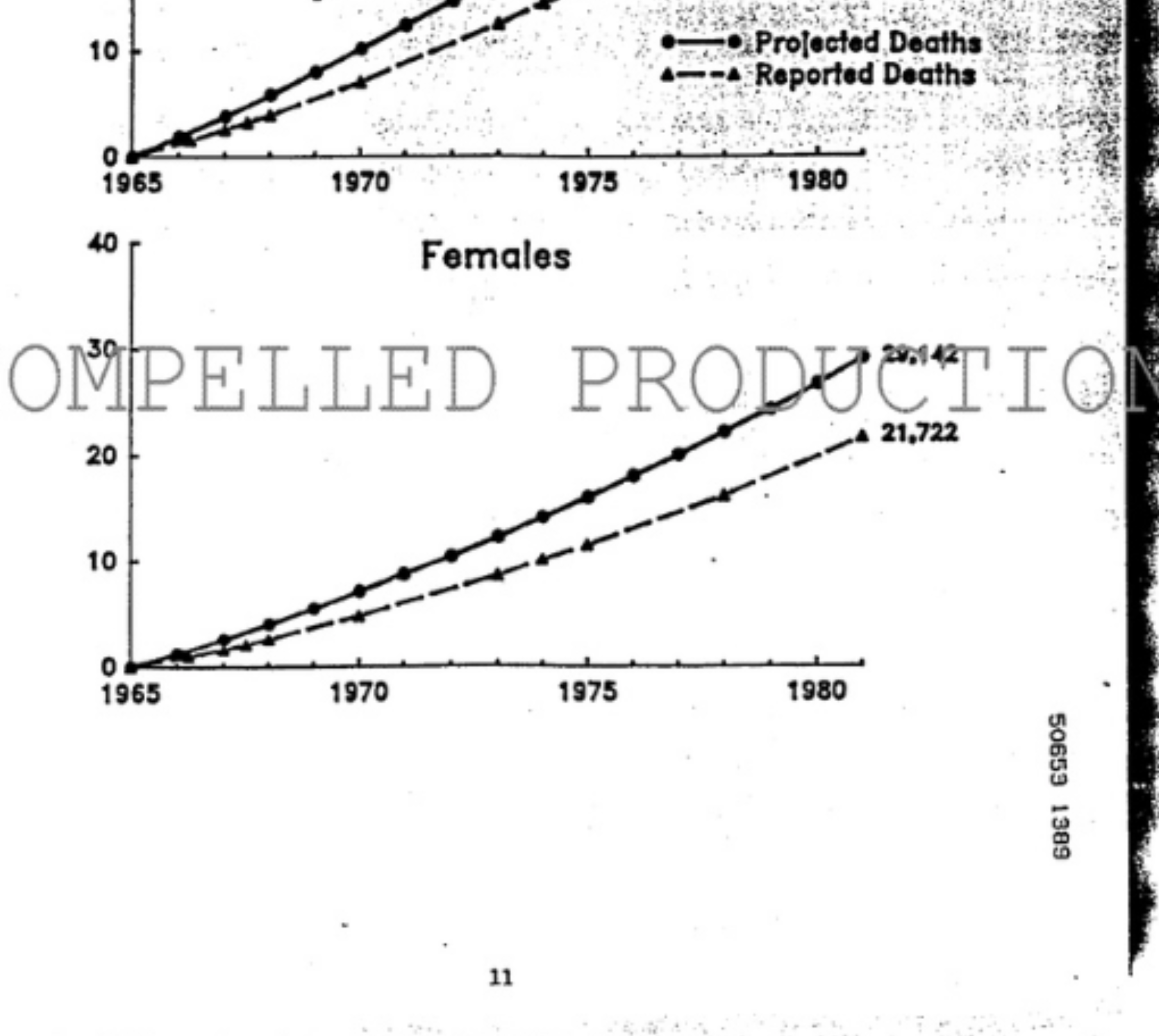}
  \caption{Original}
  \label{fig:sub1}
\end{subfigure}%
\begin{subfigure}{.5\textwidth}
  \centering
  \includegraphics[width=.6\linewidth]{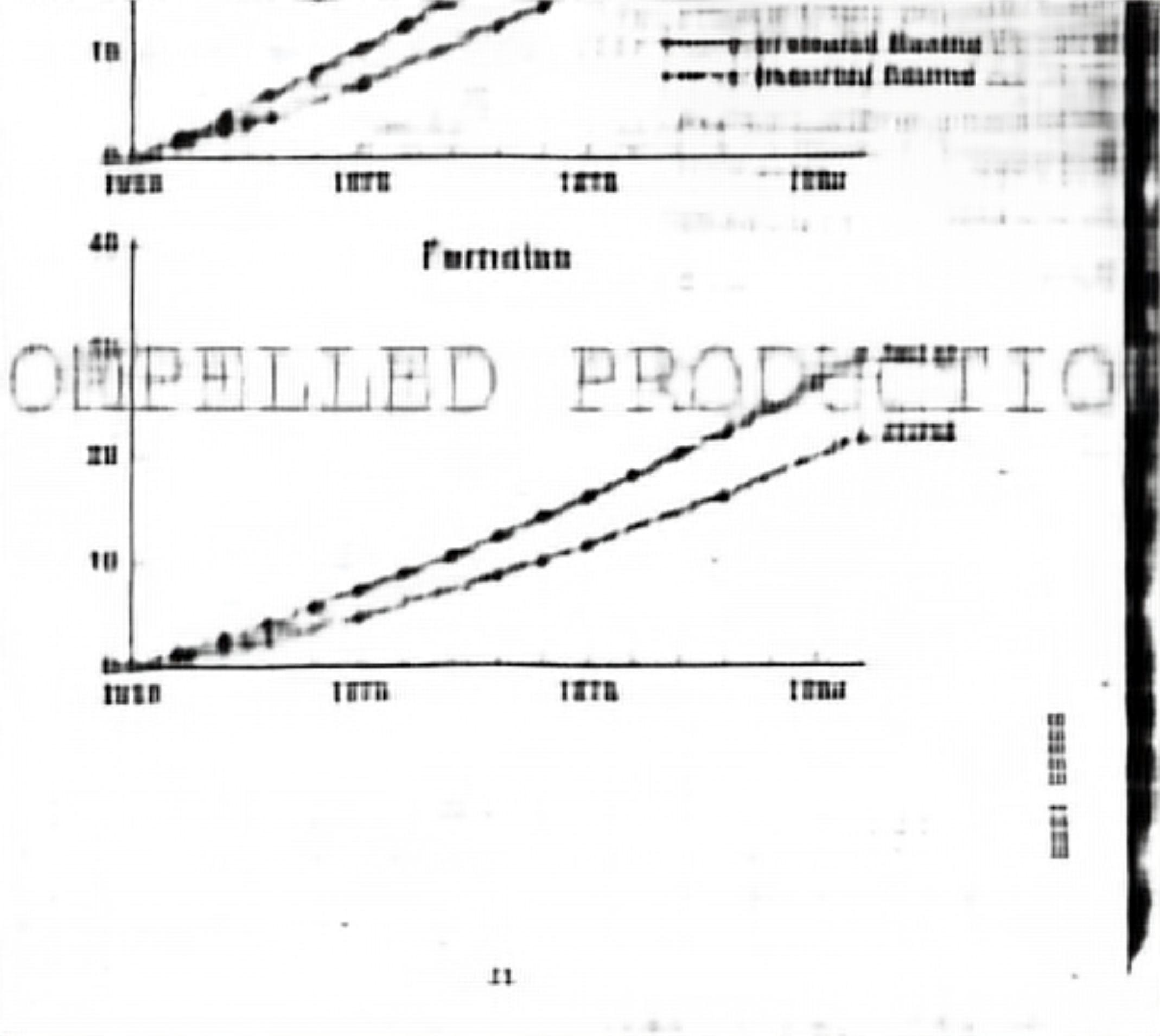}
  \caption{Reconstructed}
  \label{fig:sub2}
\end{subfigure}
\caption{Example for the Learn To Reconstruct task output on the IIT-CDIP dataset}
\label{fig:test}
\end{figure*}

\begin{align}
    \text{SmoothL1} = \begin{cases}
    0.5 (\hat{y}_n - y_n)^2,  & \text{if } |\hat{y}_n - y_n| < 1.0\\
    |\hat{y}_n - y_n| - 0.5,  & \text{otherwise}
    \end{cases}
\end{align}

\subsubsection{Text Describe Image}

In this task, we create mismatched image and text modality pairs and introduce them into our dataset with a 20\% sampling rate. The encoder is then tasked with identifying mismatched pairs. We add this task because requires a global understanding of the document as opposed to the other pre-training tasks which rely on local features. We use a cross-entropy loss denoted as $L_{TDI}$.

\subsubsection{Line Redaction}

In this task, introduced by Xu et al.\cite{DBLP:journals/corr/abs-2012-14740} we redact randomly selected lines from the input image by setting the pixels in the region to 0. The encoder is then tasked with classifying whether a given token was redacted in the input image. With this task, we seek to instill a strong sense of the spatial relationship between text and image. We use a cross-entropy loss denoted as $L_{LRed}$. \\
\\
\\
The final loss for pre-training is as follow:
\vspace{0.1pt}
\begin{align}
\begin{split}
    L = \alpha_1L_{LReg} + \alpha_2L_{LRed} +\alpha_3L_{MM-MLM} \\ 
    + \alpha_4L_{TS} + \alpha_5L_{LTR} + \alpha_6L_{TDI}
\end{split}
\end{align}

With $\alpha_{1..n} = \left\{6.0, 0.5, 1.0, 1.0, 1.0, 1.0\right\}$. The coefficient prevent early divergence by over-weighting line regression and under-weighting line redaction. 

\section{Experiments}

\subsection{Pre-training Experimental Setup}

We used a subset of the IIT-CDIP document collection\cite{DBLP:conf/sigir/LewisAAFGH06} containing 5 million scanned pages for pre-training for which we extracted the text, word-level, and line-level bounding boxes using Textract OCR.
\\
For the visual feature extraction backbone we use a pre-trained Twins\cite{DBLP:journals/corr/abs-2104-13840} model. Through their experiments, they showed that vision transformers trained with spatial attention can outperform CNN-based approaches in various dense detection tasks. We use the pre-trained Twins SVT base model and allow backpropagation to update the weights during training.\\

\subsection{Entity Extraction Tasks}

We evaluate our approach on two different datasets.

\subsubsection{CORD}

COnsolidated Receipt Dataset for Post-OCR Parsing\cite{park2019cord} is a receipt dataset containing a total of 1000 samples with 800 for training, 100 for validation, and 100 for test. It defines 30 classes grouped in 5 superclasses. F1 score is reported on the 30 classes classification task.

\subsubsection{FUNSD}

Form Understanding in Noisy Scanned Documents\cite{jaume2019} is a form dataset with a total of 199 samples, making it much smaller than other datasets we evaluated. 149 samples are used for training and the reported F1-score is reported on the remaining 50. We evaluate entire documents and do not truncate them to the first 512 sub-word tokens.\\
\\
We report our results in Table 1. We achieve a competitive F1 score on CORD, falling short of the previous SOTA by 0.94 with MATrIX which has about 3 times fewer parameters than LayoutLMv2-large and DocFormer-large. We do not achieve a high F1 score on FUNSD, which we attribute to its smaller number of samples.

\begin{table}[h!]
\centering
\setlength{\tabcolsep}{5pt}
\begin{tabular}{l c c c c c} 
 \hline
 Model & \#param (M) & FUNSD & CORD \\
 \hline
 LayoutLMv1-base & 160 & 79.27 & - \\
 LayoutLMv1-large & 390 & 77.89 & 94.93 \\
 LayoutLMv2-base & 200 & 82.76 & 94.95 \\ 
 TILT-base & 230 & - & 95.11 \\
 LayoutLMv2-large & 426 & 84.20 & 96.01 \\
 TILT-large & 780 & - & 96.33 \\
 DocFormer-base & 183 & 83.34 & 96.33 \\
 DocFormer-large & 533 & \textbf{84.55} & \textbf{96.99} \\
 \hline
 MATrIX (ours) & 166 & 78.60 & 96.05 & \\
 \hline
\end{tabular}
\caption{Entity-level F1 scores of two entity extraction tasks: FUNSD and CORD.}
\label{table:1}
\end{table}

\subsection{Document Classification Task}

This task consists in predicting a single label or class for a document page. RVL-CDIP\cite{harley2015icdar} is a dataset containing 400,000 images equally split across 16 classes. 320,000 samples are used for training, with the remaining 80,000 being equally split between the validation and test sets. The classification accuracy results are computed on the test set. Following prior work\cite{DBLP:journals/corr/abs-2106-11539}\cite{DBLP:journals/corr/abs-2012-14740}\cite{DBLP:journals/corr/abs-2108-04539}, text and spatial information is extracted using Textract OCR. We do not filter on word count and evaluate the entire test set.

We report our results in Table 2.

\begin{table}[H]
\centering
\setlength{\tabcolsep}{5pt}
\begin{tabular}{l c c} 
 \hline
 Model & \#param (M) & Accuracy \\
 \hline
 TILT-base & 230 & 93.50 \\
 TILT-large & 780 & 94.02 \\
 LayoutLMv1-base & 160 & 94.42 \\
 LayoutLMv1-large & 390 & 94.43 \\
 LayoutLMv2-base & 200 & 95.25 \\ 
 LayoutLMv2-large & 426 & 95.65 \\
 DocFormer-base & 183 & \textbf{96.17} \\
 DocFormer-large & 533 & 95.50 \\
 \hline
 MATrIX (ours) & 166 & 94.20 \\
 \hline
\end{tabular}
\caption{Classification accuracy on the RVL-CDIP dataset. For brevity we only compare against multi-modal approaches}
\label{table:1}
\end{table}

\subsection{Ablation Study}

We conduct an extensive ablation study using the CORD dataset.

\subsubsection{Impact of modality-aware relative attention}

We conduct an ablation study to determine the impact of using pre-trained BERT weights for the attention layer and sub-word token embeddings, and modality-aware relative attention on the final results for the CORD downstream task. This shows that modality-aware relative attention offers a significant improvement over regular multi-modal self-attention.

\begin{table}[h]
\centering
\setlength{\tabcolsep}{5pt}
\begin{tabular}{l c} 
 \hline
 Approach & CORD (F1) \\
 \hline
 Base & 95.05 \\ 
 Base + BERT & 95.19 \textcolor{ao(english)}{(+0.14)} \\
 Base + MATrIX & 95.48 \textcolor{ao(english)}{(+0.43)} \\
 Base + BERT + MATrIX & 96.05 \textcolor{ao(english)}{(+1.00)} \\
 \hline
\end{tabular}
\caption{Impact of the pre-training tasks on two downstream tasks' F1 score}
\label{table:1}
\end{table}
\FloatBarrier

\subsubsection{Impact of pre-training tasks}

We conduct an ablation study to determine the impact of each pre-training task on the final results for the CORD downstream task. To minimize resource usage, these pre-trainings only ran for a single epoch on the 5M dataset. In table 4, MM-MLM was always trained with the token switch task to prevent collapse. Appalaraju et al.\cite{DBLP:journals/corr/abs-2106-11539} showed that the learn to reconstruct and text describe image tasks were beneficial for this task, therefore we attribute this regression to insufficient training.

\begin{table}[h]
\centering
\setlength{\tabcolsep}{5pt}
\begin{tabular}{l c} 
 \hline
 Pre-training task & CORD (F1) \\
 \hline
 MM-MLM* & 94.52 \\ 
 + Learn to Reconstruct + Describe Image & 93.83 \textcolor{red}{(-0.69)} \\
 + Line Regression & 95.55 \textcolor{ao(english)}{(+1.72)} \\
 + Line Redaction & 95.31 \textcolor{red}{(-0.24)} \\
 \hline
\end{tabular}
\caption{Impact of the pre-training tasks on CORD F1 score}
\label{table:1}
\end{table}
\FloatBarrier

\subsubsection{Impact of modalities}

We conduct an ablation study to analyze the impact of each modality on the final results for each downstream task. We do not do additional pre-training and simply input 0 values for the missing modalities during the fine-tuning step on CORD. This shows that the additional modalities are being leveraged.

\FloatBarrier
\begin{table}[H]
\centering
\caption{Impact of the modalities on CORD F1 score}
\label{table:1}
\setlength{\tabcolsep}{5pt}
\begin{tabular}{l c c} 
 \hline
 Modality & CORD (F1) &  \\
 \hline
 Text-only & 82.18 \\
 + spatial relative attention bias & 93.80 \textcolor{ao(english)}{(+11.62)} \\
 + spatial features embeddings & 95.38 \textcolor{ao(english)}{(+1.58)} \\
 + visual features & 96.05 \textcolor{ao(english)}{(+0.67)} \\
 \hline
\end{tabular}
\end{table}

\subsubsection{Impact of input resolution}

We conduct an ablation study to measure the impact of increasing and decreasing the input resolution when fine-tuning, given an encoder pre-trained at a maximum resolution of 512x512.
We observe indeed that the model is resilient to resolution change.

\FloatBarrier
\begin{table}[H]
\centering
\caption{Impact of the input resolution on CORD F1 score}
\label{table:1}
\setlength{\tabcolsep}{5pt}
\begin{tabular}{l l c} 
 \hline
 Input resolution & Maximum visual tokens & CORD (F1) \\
 \hline
 256 & 64 (8x8) & 95.83 \\ 
 512 & 256 (16x16) & 96.05 \\
 768 & 576 (24x24) & 96.05 \\
 \hline
\end{tabular}
\end{table}

\section{Conclusion}

In this work, we introduced MATrIX, a multi-modal transformer for information extraction in the visual document understanding field. We presented a novel yet simple modality-aware relative attention mechanism which we demonstrated to be superior to the simpler relative attention and reached competitive results on three datasets. Additionally, we showed that using word-level tokens is a practical way of working with longer text sequences. We introduced line regression and token switch, two new cross-modality tasks which were shown to improve F1 score on the CORD dataset.\\ In future work, we will explore MATrIX suitability as a backbone for more complex tasks such as table detection, table structure extraction, and entity-relationship prediction. 

\section{Supplemental}

\subsection{Implementation details}

The pre-training was done on 8x V100 with 32GB of memory using the LAMB\cite{DBLP:journals/corr/abs-1904-00962} algorithm which was shown to decrease the training time of transformer architecture significantly. We used the implementation available in the Nvidia Apex library.
The fine-tuning for the FUNSD and CORD downstream tasks were done on a single V100 with 16GB of memory.

\begin{table*}[h!]
\centering
\setlength{\tabcolsep}{5pt}
\begin{tabular}{l | c | c | c | c } 
 \hline
 \multirow{2}{*}{Hyper-parameter} & \multirow{2}{*}{Pre-training} & \multicolumn{3}{c}{Fine-tuning} \\
 \cline{3-5}
 & & CORD & FUNSD & RVL-CDIP \\ 
 \hline
 Epochs & 5 & 100 & 100 & 5 \\ 
 Learning rate & 0.0001 & 0.0001 & 0.00005 & 0.0001 \\
 Warm-up & 60 000 & 1000 & 1000 & 60 000\\
 Gradient Clipping & 2.0 & 4.0 & 4.0 & 2.0 \\
 Optimizer & FusedLAMB & AdamW & AdamW & AdamW\\
 Lower case & Yes & Yes & Yes & Yes\\
 Sequence length & 512 & 512 & 512 & 512\\
 Encoder layers & 12 & 12 & 12 & 12\\
 Batch size & 8 & 4 & 1 & 8\\
 GPU & V100 (32GB) & V100 (16GB) & V100 (16GB) & V100 (32GB)\\
 \hline
\end{tabular}
\caption{Hyper-parameters used for pre-training and fine-tuning on the various tasks}
\label{table:1}
\end{table*}

\clearpage
\bibliographystyle{ieee_fullname}
\bibliography{egbib}
\end{document}